\documentclass[journal,onecolumn]{IEEEtran}

\usepackage{amsmath}
\usepackage{amssymb}
\usepackage{amsthm}
\usepackage{color}
\usepackage{array}
\usepackage{cite}
\usepackage{enumerate}
\usepackage{graphicx}
\usepackage[hang]{subfigure}
\usepackage{footmisc}
\usepackage{amsbsy}
\usepackage{bm}
\usepackage{cases}
\usepackage{multirow}
\usepackage{algorithmic}
\usepackage{comment}
\usepackage{times}
\usepackage{paralist}
\usepackage[ruled]{algorithm2e}
\usepackage{setspace}
\usepackage[breakable]{tcolorbox}
\usepackage{kotex}
\usepackage{url}

\renewcommand{\baselinestretch}{1.05}

\usepackage{bm}

\begin{document}
\title{End-to-End Learning of Neuromorphic Wireless Systems for Low-Power Edge Artificial Intelligence}

\author{Nicolas~Skatchkovsky,~\IEEEmembership{Student Member,~IEEE,}
        Hyeryung~Jang,~\IEEEmembership{Member,~IEEE,}
        and~Osvaldo~Simeone,~\IEEEmembership{Fellow,~IEEE}% <-this % stops a space
\thanks{The authors are with King's College Communication, Learning and Information Processing lab, Centre for Telecommunications Research, Department of Engineering, King's College London, UK. The authors have received funding from the European Research Council (ERC) under the European Union’s Horizon 2020 Research and Innovation Programme (Grant Agreement No. 725731).}}

% make the title area
\maketitle

% As a general rule, do not put math, special symbols or citations
% in the abstract or keywords.
\begin{abstract}
This paper introduces a novel ``all-spike'' low-power solution for remote wireless inference that is based on neuromorphic sensing, Impulse Radio (IR), and Spiking Neural Networks (SNNs). In the proposed system, event-driven neuromorphic sensors produce asynchronous time-encoded data streams that are encoded by an SNN, whose output spiking signals are pulse modulated via IR and transmitted over general frequence-selective channels; while the receiver's inputs are obtained via hard detection of the received signals and fed to an SNN for classification. We introduce an end-to-end training procedure that treats the cascade of encoder, channel, and decoder as a probabilistic SNN-based autoencoder that implements Joint Source-Channel Coding (JSCC). The proposed system, termed NeuroJSCC, is compared to conventional synchronous frame-based and uncoded transmissions in terms of latency and accuracy. The experiments confirm that the proposed end-to-end neuromorphic edge architecture provides a promising framework for efficient and low-latency remote sensing, communication, and inference.
\end{abstract}

% Note that keywords are not normally used for peerreview papers.
\IEEEpeerreviewmaketitle

\section{Introduction}
\subsection{Motivation}
\label{sec:motivation}
\IEEEPARstart{I}{nternet} of Things (IoT) networks are cyberphysical systems that carry out sensing, processing, learning, and communication. An important class of IoT networks is given by edge-based systems that target the continuous sensing and processing of video, radio, audio, or other types of physical signals, on battery-powered mobile devices for sensing and edge servers for inference. Examples of use cases include mobile personal healthcare – data collected by wearables, such as ECG signals, are processed at a smartphone in order to detect anomalies; and drone-based monitoring – video signals captured by unmanned aerial vehicles are analysed at mobile ground terminals. In these systems, energy efficiency is typically of paramount importance, and deploying a network of conventional “always-on” IoT devices that stream data through digital sensors, processors (CPUs), and Radio Frequency (RF) transceivers, with their associated clocks, mixers, analog-to-digital and digital-to-analog converters, and phase locked loops, may not be feasible \cite{neuromorhpic_efficiency, kaur2017efficiency}. 

\subsection{End-to-End Neuromorphic Wireless Systems}
\label{sec:review}
\begin{figure}
    \centering
    \includegraphics[scale=0.3]{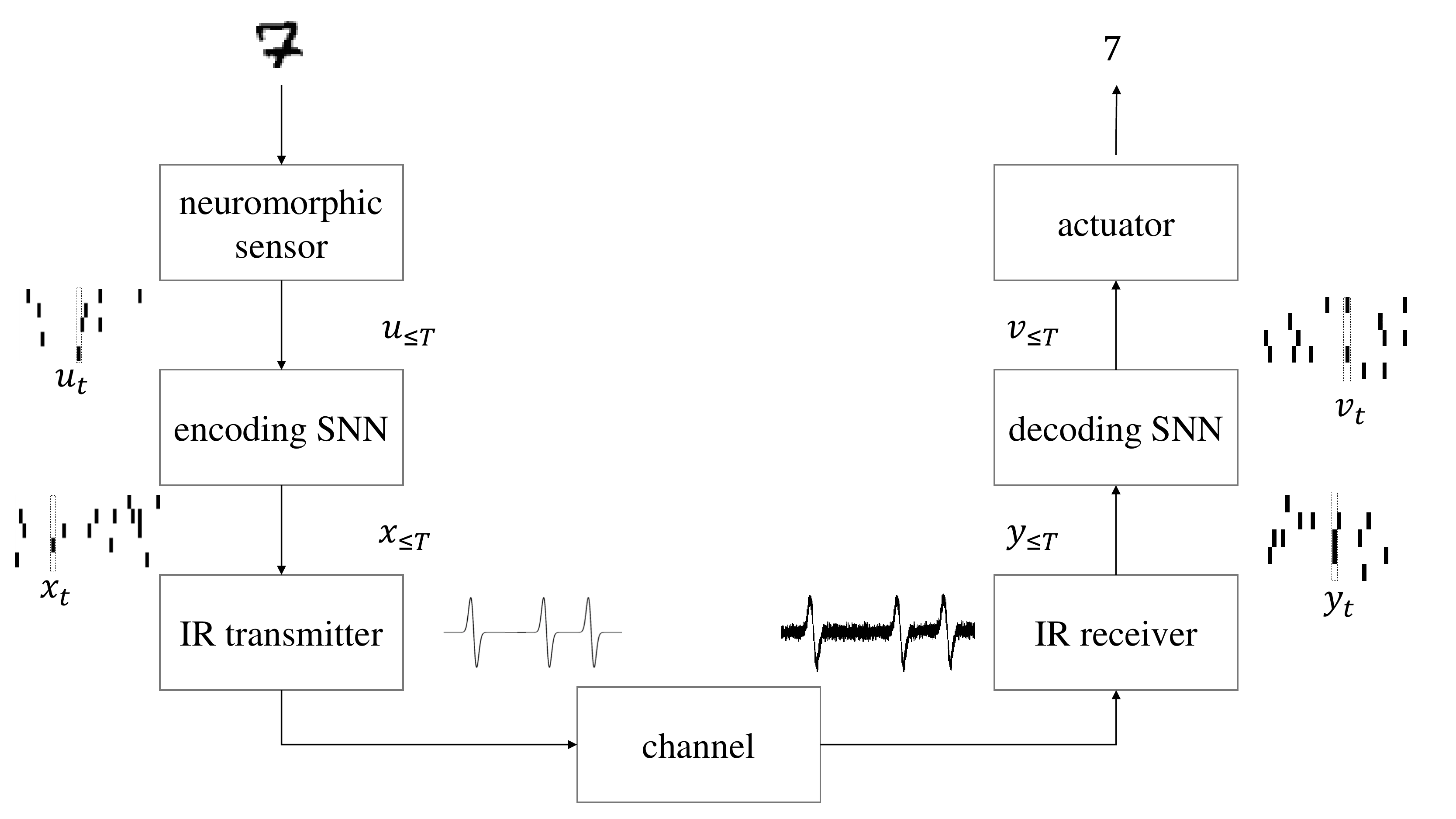}
    \caption{NeuroJSCC: An end-to-end spike-domain remote inference system that implements Joint Source-Channel Coding based on neuromorphic sensing, Spiking Neural Networks (SNNs), and Impulse Radio (IR).}
    \label{system}
\end{figure}

To tackle the inefficiency of conventional IoT mobile edge architectures, in this paper, we propose the architecture illustrated in Fig.~\ref{system} that replaces each of the three digital blocks – sensor, CPU, and transmitter/receiver - with \textit{neuromorphic} counterparts \cite{mead1990neuromorphic, rajendran2019neuromorphic}. Inspired by the workings of biological brains, these blocks consume energy in an \textit{event-driven} and \textit{sparse} fashion, by leveraging encoding and signalling based on temporal spikes. 

In \textit{neuromorphic sensors}, spikes mark the occurrence of a relevant event, e.g., a significant change in a pixel for a neuromorphic camera \cite{Lichtsteiner2006dvscamera}. Extremely low energy is consumed when the monitored scene is idle. In neuromorphic processors, known as \textit{Spiking Neural Networks (SNNs)}, spiking signals are processed via dynamic neural models for the detection of spatio-temporal patterns. SNNs have recently emerged as a biologically plausible alternative to Artificial Neural Networks (ANNs) \cite{maass1997snn}, with significant benefits in terms of energy efficiency and latency \cite{davies2019spikemark}. Finally, for communications, pulses, or spikes, can encode information for radio signalling via low-power \textit{Impulse Radio (IR)}. Commercial products are available for all these blocks, including DVS cameras \cite{Lichtsteiner2006dvscamera}, Intel's Loihi SNN chip \cite{davies2018loihi}, and transceivers implementing the IEEE $802.15.4$z IR standard \cite{impulse_radio}. Beyond-5G systems based on terahertz communications may also leverage IR \cite{yu2015thzir}.

As seen in Fig.~\ref{system}, the proposed system consists of the integration of neuromorphic sensing and processing with IR transmission, and carries out Joint Source-Channel Coding (JSCC), as it performs source and channel coding in a single step. The key advantages of such an ``all-spike'' neuromorphic solution are latency and energy efficiency: since spikes are produced by the neuromorphic sensor only in the presence of changes in the monitored process, the system consumes negligible energy when no relevant information is recorded, and it responds immediately to significant events. For all blocks, the energy spent is essentially proportional to the number of spikes processed, with the energy per spike being as low as a few picojoules in the case of SNNs \cite{intro_snn} and IR \cite{wolf2011lowpowerir}. This is unlike conventional frame-based sytems in which frames are produced periodically -- entailing a generally delayed response -- and continuously -- causing energy consumption. 

Following a recent line of research \cite{oshea2017intro, simeone2018intro}, \cite{choi2019jssc, raj2019jssc}, we propose to train the system in Fig.~\ref{system} in an end-to-end manner. We specifically view the overall system as an autoencoder under the assumption that the channel model is known. Furthermore, in order to overcome the difficulties due to the non-differentiability of the activation function of deterministic SNN models \cite{neftci2019surrogate}, we focus on a probabilistic implementation of SNNs \cite{intro_snn}. Alternative solutions include the methods in \cite{neftci2019surrogate, huh2018gradient, zenke2018superspike, neftci2017eventbackprop}.

\subsection{Related Work}
\label{sec:related}
Neuromorphic sensors are commercially available and have been deployed in a number of real-world applications, including for video surveillance, hearing aids, biomedical signal processing, and Lidar \cite{schuman2017survey, vanarse2016review}. IR has been proposed as a means to communicate wirelessly digital packets between SNN chips \cite{cassidy2008ir}; and to transmit time-encoded analog signals, akin to those measured by neuromorphic sensors, for biomedical applications \cite{shahshahani2015ir}. General public IR applications include Apple's U1 TMKA75 Ultra Wideband chip in the latest versions of their smartphones \cite{appleiphone11}. A combination of neuromorphic sensing, time-based computing, and IR has been used in \cite{pepr2016sensor} to implement a consensus method based on device-to-device local communications for the purpose of computing the maximum of scalar observations. This work is, to the best of our knowledge, the first attempt at considering an end-to-end neuromorphic system where sensing, communication, and processing is natively performed in the spike domain, while taking into account the key problems of training SNNs and wireless channel impairments.

Recent work has demonstrated the potential of joint source-channel coding based on conventional (non-spiking) neural networks. With the aim of transmitting images over wireless channels, reference \cite{bourtsoulatze2019jssc} jointly trains encoding and decoding convolutional neural networks (CNNs) as an autoencoder with the channel as a non-trainable middle layer under a mean squared error loss function. More recent works adopt information-theoretic criteria, such has the maximization of the mutual information between the input data and received noisy codeword \cite{choi2019jssc, ullrich2020jssc}, or of the expected marginal likelihood of data points at the output of the decoder \cite{raj2019jssc}. 

\section{System Model}
\subsection{Overview}
\label{sec:overview}
As illustrated in Fig.~\ref{system}, we consider an edge IoT link aimed at communicating time-encoded information, e.g., sensed by a DVS camera, to a receiver for the purpose of remote inference. In order to leverage the efficiency of neuromorphic computing and IR, we assume that encoding, transmission, and decoding are performed using spiking signals, with the end of performing a supervised learning task at the receiver. 

The signal sensed by the neuromorphic sensor, e.g., a DVS camera, is encoded as a vector of time samples $\mathbf{u}_{\leq T} = (\mathbf{u}_1, \ldots, \mathbf{u}_T)$ across $T$ timesteps. At each time $t=1, \ldots, T$, observation $\mathbf{u}_t = [u_{1,t},\ldots,u_{d_u,t}]$ is a $d_u \times 1$ vector of binary, or spiking, signals $u_{i,t} \in \{0,1\}$. This observation is recorded in conditions that identify a given class, e.g., a hand-written digit or a gesture displayed to the camera. Let the desired output of the decoding SNN be given by a spiking signal $\mathbf{v}_{\leq T} = (\mathbf{v}_1,\ldots,\mathbf{v}_T)$, with $\mathbf{v}_t \in \{0,1\}^{d_v}$ for $t=1,\ldots,T$. Inference can be carried out by decoding the output signal $\mathbf{v}_{\leq T}$ to a class index using standard methods for SNN-based classification \cite{intro_snn}, \cite{stanojevi2020FileCB}. For example, rate decoding predicts a class by selecting the neurons in an output layer with the largest number of spikes. The joint distribution $p(\mathbf{u}_{\leq T}, \mathbf{v}_{\leq T})$ of input and desired output is unknown, and, as in standard supervised learning formulations, a data set of pairs $\{ (\mathbf{u}_{\leq T}, \mathbf{v}_{\leq T}) \}$ is available for training.

The encoding SNN is a causal mapping that takes the spiking signals $\mathbf{u}_{\leq T}$ as input and outputs binary signals $\mathbf{x}_{\leq T} = (\mathbf{x}_1,\ldots,\mathbf{x}_T)$ with $\mathbf{x}_t = \{0,1\}^{d_x}$ for each time $t$. We define $r = d_x / d_u$ as the \textit{rate} of the communication scheme. The signals $\mathbf{x}_{\leq T}$ are modulated using $T$ symbol periods of $d_x$ parallel IR transmissions, with each spike encoded by an IR waveform such as a Gaussian monopulse. As we will detail in Sec.~\ref{sec:snn_model}, the encoding SNN defines a probabilistic mapping $p_{\theta^E}(\mathbf{x}_{\leq T} || \mathbf{u}_{\leq T})$, where $\theta^E$ is a learnable parameter vector of the SNN. Here, and throughout the paper, we use the \textit{causally conditional} notation \cite{kramer1998directed} $p_{\theta^{E}}(\mathbf{x}_{\leq T} ||\mathbf{u}_{\leq T}) = \prod_{t=1}^{T} p_{\theta^{E}}(\mathbf{x}_{t} | \mathbf{x}_{\leq t - 1}, \mathbf{u}_{\leq t})$ to denote the causal dependency of the output $\mathbf{x}_{\leq T}$ on the input $\mathbf{u}_{\leq T}$.

The channel is modeled as a stochastic system with memory whose output $\mathbf{y}_{t}$ at each time $t = 1, \dots, T$ follows a probability distribution conditional on its past inputs $\mathbf{x}_{\leq t-1}$ up to time $t - 1$. This can be expressed as the causally conditional distribution
\begin{align}
p(\mathbf{y}_{\leq T} || \mathbf{x}_{\leq T}) = \prod_{t=1}^{T} p(\mathbf{y}_{t} | \mathbf{y}_{\leq t - 1}, \mathbf{x}_{\leq t}).    
\end{align}
The channel $p(\mathbf{y}_{t} | \mathbf{y}_{\leq t - 1},  \mathbf{x}_{\leq t})$ includes the effect of waveform generation at the transmitter, intersymbol interference, as well as filtering and sampling at the receiver. As a result, we assume that the received signal $\mathbf{y}_{t} \in \{0, 1\}^{d_y}$ is binary. For example, a frequency-flat Gaussian channel with noise power $\sigma^{2}$ and threshold decoding can be modelled as
\begin{align}
    \label{eq:channel}
    \mathbf{y}_{t} = Q(\mathbf{x}_{t} + \mathbf{n}_{t}), 
\end{align}
where $\mathbf{n} \sim \mathcal{N}(0, \sigma^{2}I_{d_x})$; $d_y = d_x$; and $Q( \cdot )$ is an element-wise binary quantizer. We assume that the channel model $p(\mathbf{y}_{t} | \mathbf{y}_{\leq t - 1}, \mathbf{x}_{\leq t})$ is known during training, so that it can be used to draw samples from the received signal given the input. 

Finally, the receiver maps the signals $\mathbf{y}_{\leq T}$ to the output $\mathbf{v}_{\leq T}$ with $v_t \in \{0, 1\}^{d_v}$, via a decoding SNN, which defines a probabilistic mapping $p_{\theta^{D}}(\mathbf{v}_{\leq T}||\mathbf{y}_{\leq T})$ with a learnable parameter vector $\theta^{D}$. Combining all elements, the joint distribution $p_{\theta}(\mathbf{u}_{\leq T}, \mathbf{v}_{\leq T})$ of the end-to-end system is parameterized by $\theta = \{ \theta^{E}, \theta^{D} \}$.

\subsection{Probabilistic SNN Model}
\label{sec:snn_model}

\begin{figure}
\centering
\includegraphics[scale=0.3]{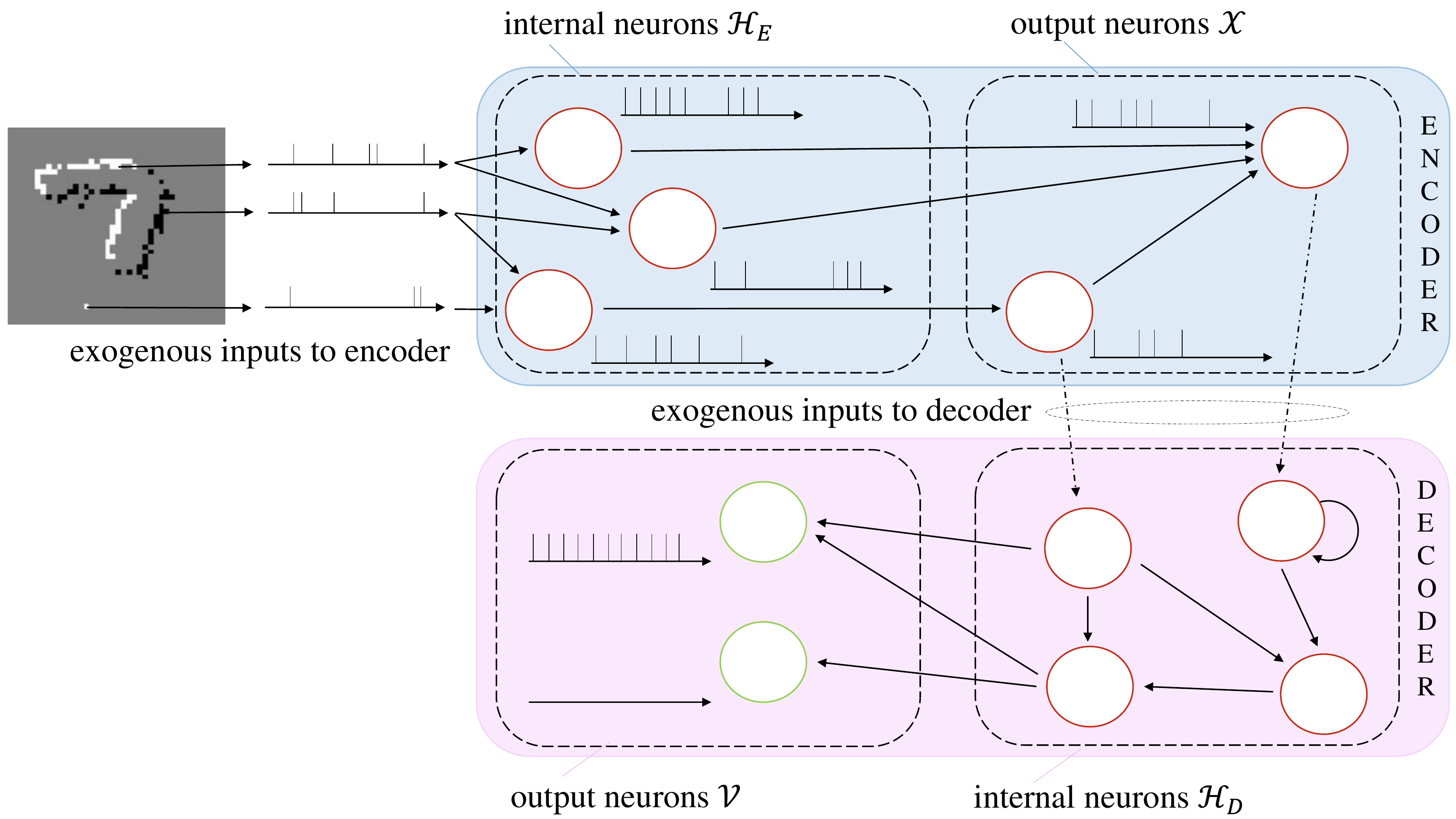}
\caption{Illustration of a SNN. The solid directed links between two neurons represent synapses, while dashed arrows describe the effect of the channel.}
\label{snnmodel}
\end{figure}

Both encoding and decoding SNNs operate according to the standard probabilistic Generalized Linear Model (GLM) \cite{intro_snn}. We use superscripts $S \in \{ E, D \} $ to indicate either the encoding (E) or decoding (D) SNN. As illustrated in Fig.~\ref{snnmodel}, each SNN consists of a set $\mathcal{N}^S$ of spiking neurons connected via an arbitrary directed graph, possibly with cycles. Each neuron $i$ receives the signals emitted by the subset $\mathcal{P}_{i}$ of neurons connected to it through directed links, known as \textit{synapses}, which we take to include also the \textit{exogeneous inputs}. At any time-step $t = 1, \dots, T$, each neuron $i$ outputs a binary signal $s_{i, t} \in \{0, 1\}$, with ``$1$'' representing a the firing of a spike. We collect in vector $\mathbf{s}_t = (s_{i,t} : i \in \mathcal{N}^S)$ the spikes emitted by all neurons $\mathcal{N}^S$ at time $t$ and denote by $\mathbf{s}_{\leq t} = (\mathbf{s}_1,\ldots,\mathbf{s}_t)$ the spike signals up to time $t$.

The conditional spiking probability of a neuron $i \in \mathcal{N}^S$ at time $t$ is defined as
\begin{align} \label{eq:spiking_proba}
    p_{\theta_i^{S}}( s_{i,t} = 1 | \mathbf{s}_{\leq t-1}) = p_{\theta_i^{S}}(s_{i,t} = 1 | o_{i,t}) = \sigma(o_{i,t}),
\end{align}
with $\sigma(\cdot)$ being the sigmoid function and $\theta_i^{S}$ representing the local model parameters of neuron $i$. In \eqref{eq:spiking_proba}, the dependency on the history $\mathbf{s}_{\leq t-1}$ is mediated by the neuron's {\em membrane potential} $o_{i,t}$. As defined below, the membrane potential is obtained as the output of spatio-temporal moving average filters with finite-duration for both synapses and self-memory.
From \eqref{eq:spiking_proba}, the spiking probability \eqref{eq:spiking_proba} increases with the membrane potential and the negative log-probability corresponds to the binary cross-entropy, i.e., 
\begin{align*} %\label{eq:binary-ind-ce}
    & - \log p_{\theta_i}(s_{i,t} | o_{i,t}) = \ell \big( s_{i,t}, \sigma(o_{i,t}) \big) \cr 
    &\quad \quad := - s_{i, t} \log\big(\sigma(o_{i,t})\big) - (1 - s_{i, t}) \log\big(1 - \sigma(o_{i,t})\big).
\end{align*}
The joint probability of the spike signals $\mathbf{s}_{\leq T}$ up to time $T$, causally conditioned on the exogeneous inputs $\mathbf{e}_{\leq T}$ is defined using the chain rule as $p_{\theta^S}(\mathbf{s}_{\leq T} || \mathbf{e}_{\leq T}) = \prod_{t=1}^T \prod_{i \in \mathcal{N}^S} p_{\theta_i^{S}}(s_{i,t} | o_{i,t})$, where $\theta^S = \{\theta_i^{S}\}_{i \in \mathcal{N}^S}$ is the vector of parameters of the SNN. 

To account for synaptic memory, as in \cite{intro_snn}, we define $K$ finite-duration filters $\{a_t^{(k)}\}_{k=1}^K$, and, for neural self-memory, we similarly introduce a finite-duration filter $b_t$ (multiple somatic filters could also be considered). Denoting by $f_t \ast g_t$ the convolution operator $f_t \ast g_t = \sum_{\delta > 0} f_\delta g_{t-\delta}$, each $(j,i)$ synapse between pre-synaptic neuron $j$ and post-synaptic neuron $i$ computes the \textit{synaptic filtered trace} 
\begin{align} %\label{eq:binary-filtered-traces}
    \overrightarrow{s}_{j,t}^{(k)} = a_t^{(k)} \ast s_{j,t},
\end{align} 
while the soma of each neuron $i$ computes the \textit{feedback, or self-memory, trace} $\overleftarrow{s}_{i,t} = b_t \ast s_{i,t}$.
The membrane potential of neuron $i$ at time $t$ is then given as the weighted sum
\begin{equation} \label{eq:binary-potential}
    o_{i,t} = \sum_{j \in \mathcal{P}_i} \sum_{k=1}^K w_{j,i}^{(k)} \overrightarrow{s}_{j,t-1}^{(k)} + w_i \overleftarrow{s}_{i,t-1} + \gamma_i,
\end{equation}
where $\{w_{j,i}^{(k)}\}_{k=1}^K$ is the set of learnable synaptic weights from pre-synaptic neuron $j \in \mathcal{P}_i$ to post-synaptic neuron $i$; $w_i$ is the learnable feedback weight; and $\gamma_i$ is a learnable bias parameter, with $\theta_{i}^{S} = \{ \{\{w_{j,i}^{(k)}\}_{k=1}^K\}_{j \in \mathcal{P}_i}, w_i, \gamma_i \}$ being the local model parameters for neuron $i$.

\section{NeuroJSCC}
\label{sec:varjscc}
In this section, we describe the proposed NeuroJSCC system, that implements JSCC in the system of Fig~\ref{system} by using two SNNs: one to encode the sensed signal $\mathbf{u}_{\leq T}$ and the other to decode and process the signal $\mathbf{y}_{\leq T}$ received through the wireless channel. The design of the two SNNs is carried out by maximizing the log-likelihood that the decoding SNN outputs desired spiking signals $\mathbf{v}_{\leq T}$ in response to a given input $\mathbf{u}_{\leq T}$. Mathematically, we wish to select the model parameters $\theta$ for all encoding and decoding neurons by addressing the problem
\begin{align}
\label{eq:mlproblem}
\min_{\theta} ~ - \log p_{\theta}(\mathbf{v}_{\leq T} || \mathbf{u}_{\leq T}).
\end{align}

As can be seen in Fig.~\ref{snnmodel}, the set $\mathcal{N}^E$ of neurons in the encoding SNN is partitioned into the subsets $\mathcal{X}$ of output and $\mathcal{H}^E$ of internal neurons; and the set $\mathcal{N}^D$ of neurons in the decoding SNN is partitioned into the subsets $\mathcal{V}$ of output and $\mathcal{H}^D$ of internal neurons. During training, the desired spiking signals $\mathbf{v}_{\leq T}$ of the output neurons in $\mathcal{V}$ are specified by the training data, i.e., $s_{i,t} = v_{i,t}$ for $i \in \mathcal{V}$; while the stochastic spiking signals $\mathbf{x}_{\leq T}, \mathbf{h}_{\leq T}^E$ and $\mathbf{h}_{\leq T}^D$ of the neurons in subsets $\mathcal{X}, \mathcal{H}^E$ and $\mathcal{H}^D$ are not observed. They should be adapted during training to ensure the desired output behavior $\mathbf{v}_{\leq T}$ of the decoding SNN.

Considering the stochastic channel discussed in Sec.~\ref{sec:overview}, the hidden spiking signals $\mathbf{h}_{\leq T}^E, \mathbf{x}_{\leq T}, \mathbf{y}_{\leq T}$ and $\mathbf{h}_{\leq T}^D$ have to be averaged over in order to evaluate the log-likelihood in \eqref{eq:mlproblem} as
\begin{align}
    \label{log_likelihood}
    & \log p_{\theta}(\mathbf{v}_{\leq T}|| \mathbf{u}_{\leq T}) & \nonumber \\
    % &= \log \sum_{\substack{\mathbf{h}^{E}_{\leq T}, \mathbf{x}_{\leq T}, \\ \mathbf{y}_{\leq T}, \mathbf{h}^{D}_{\leq T}}} p_{\theta}(\mathbf{v}_{\leq T}, \mathbf{h}^{E}_{\leq T}, \mathbf{x}_{\leq T}, \mathbf{y}_{\leq T}, \mathbf{h}^{D}_{\leq T} || \mathbf{u}_{\leq T}),
    % &= \log \mathrm{E}_{p_{\theta^E}(\mathbf{h}_{\leq T}^E, \mathbf{x}_{\leq T} || \mathbf{u}_{\leq T}) p(\mathbf{y}_{\leq T} || \mathbf{x}_{\leq T}) p_{\theta^D}( \mathbf{h}_{\leq T}^D || \mathbf{y}_{\leq T})} p_{\theta^{D}}(\mathbf{v}_{\leq T} || \mathbf{y}_{\leq T}, \mathbf{h}^{D}_{\leq T}),
    &= \log \mathrm{E}_{p_{\theta}(\mathbf{h}_{\leq T}^E, \mathbf{x}_{\leq T}, \mathbf{y}_{\leq T},  \mathbf{h}_{\leq T}^D  || \mathbf{u}_{\leq T})} \Big[ p_{\theta^{D}}(\mathbf{v}_{\leq T} || \mathbf{y}_{\leq T}, \mathbf{h}^{D}_{\leq T}) \Big],
\end{align}
since the joint distribution of all variables of interest factorizes as
\begin{align*}
    & p_{\theta}(\mathbf{h}_{\leq T}^E, \mathbf{x}_{\leq T}, \mathbf{y}_{\leq T}, \mathbf{h}_{\leq T}^D || \mathbf{u}_{\leq T})p_{\theta^D}(\mathbf{v}_{\leq T}|| \mathbf{y}_{\leq T}, \mathbf{h}_{\leq T}^D ) \cr
    & = p_{\theta^E}(\mathbf{h}_{\leq T}^E, \mathbf{x}_{\leq T} || \mathbf{u}_{\leq T}) p(\mathbf{y}_{\leq T} || \mathbf{x}_{\leq T}) p_{\theta^D}(\mathbf{v}_{\leq T}, \mathbf{h}_{\leq T}^D || \mathbf{y}_{\leq T}),
\end{align*}
with $p(\mathbf{y}_{\leq T} || \mathbf{u}_{\leq T}, \mathbf{x}_{\leq T})$ being the stochastic channel, and $p_{\theta^E}(\mathbf{h}_{\leq T}^E, \mathbf{x}_{\leq T} || \mathbf{u}_{\leq T})$ and $p_{\theta^D}(\mathbf{v}_{\leq T}, \mathbf{h}_{\leq T}^D || \mathbf{y}_{\leq T})$ being the joint distribution defined by the encoding and decoding SNN, respectively. 

%Via maximum likelihood (ML) training, we wish to select the model parameters of the encoding and decoding SNNs $\theta_{E}$ and $\theta_{D}$ such as to obtain the desired input/output behavior of the system, while appropriately modulating the operations of other neurons, namely the internal neurons of the encoding and decoding SNNs in the sets $\mathcal{H}_{E}$ and $\mathcal{H}_{D}$, and the output neurons of the encoding SNN in the set $\mathcal{X}$.

\begin{figure}
\centering
\includegraphics[scale=0.31]{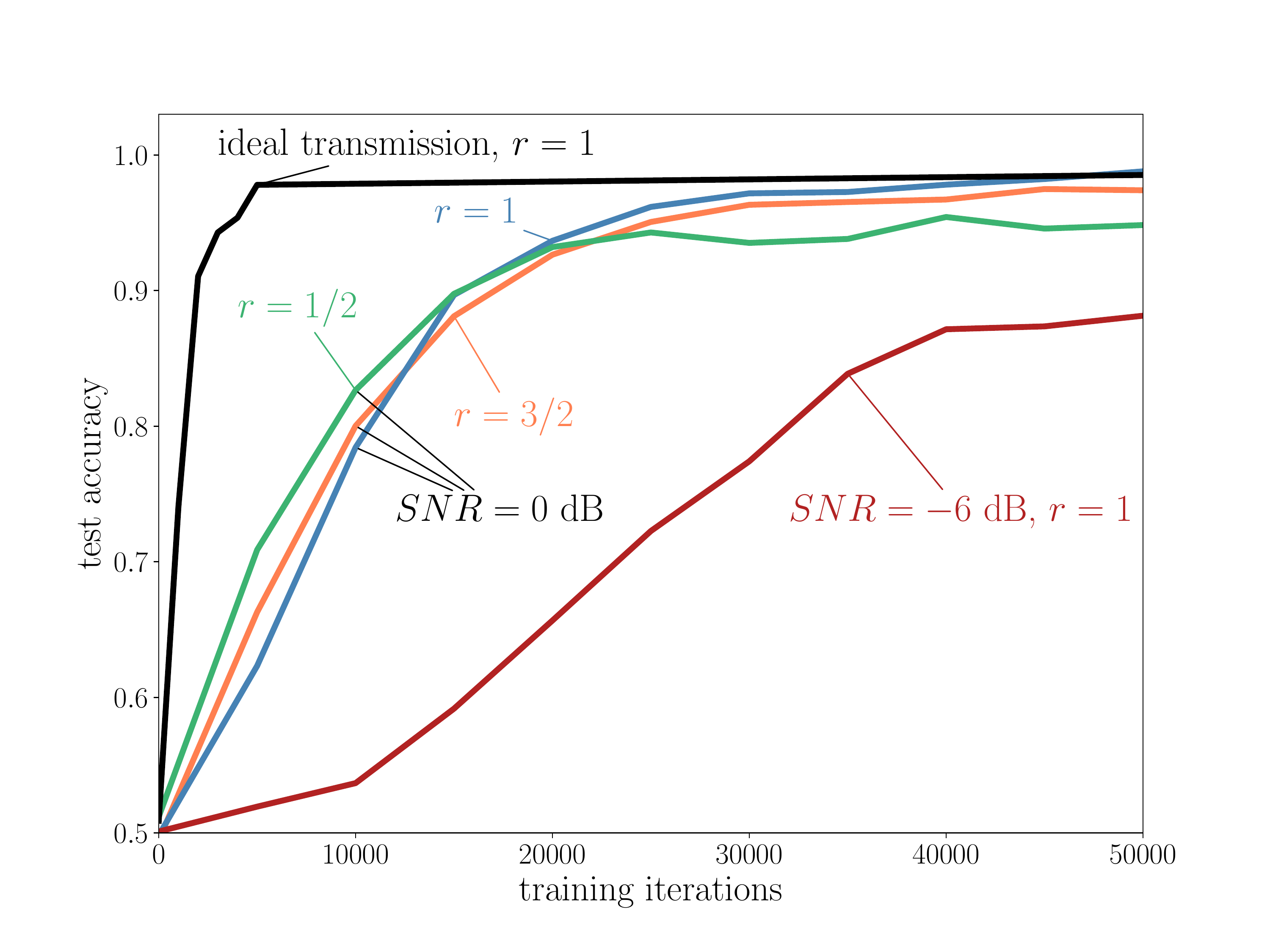}
\caption{Test accuracy of NeuroJSCC as a function of the number of training iteration as compared to the case of ideal transmission (i.e., of a decoding SNN having direct access to the data).}
\label{fig:acc_per_train_ite}
\end{figure}

To address problem \eqref{eq:mlproblem}, we minimize the upper bound obtained from Jensen's equality (see e.g., \cite[Ch.~6~and~Ch.~8]{simeone2018intro} for more details)
\begin{align} \label{elbo}
    - \log p_{\theta}(\mathbf{v}_{\leq T} || \mathbf{u}_{\leq T}) & \leq \mathcal{L}(\theta)\cr 
    & := \mathrm{E}_{\theta}\Big[- \log p_{\theta^{D}}(\mathbf{v}_{\leq T}||\mathbf{y}_{\leq T}, \mathbf{h}_{\leq T}^D ) \Big] \cr
    & = \sum_{t=1}^T \mathrm{E}_{\theta}\Big[ - \log p_{\theta^D}(\mathbf{v}_t || \mathbf{y}_{\leq t} \mathbf{h}^{D}_{\leq t}) \Big] \cr 
    & = \sum_{t=1}^T \sum_{i \in \mathcal{N}} \mathrm{E}_{\theta}\Big[\ell \big( v_{i,t}, \sigma(o_{i,t}) \big) \Big] \cr 
    & := \mathrm{E}_{\theta}\Big[\ell_{\theta} \Big],
\end{align}
where the expectation is as in \eqref{log_likelihood}.

Finally, we tackle the maximization of the lower bound $\mathcal{L}(\theta)$ via gradient descent, where the gradient is computed as \cite[Ch.~8]{simeone2018intro}
\begin{align} \label{eq:grad-elbo}
    &\nabla_{\theta} \mathcal{L}(\theta)
    % & = \mathrm{E}_{\theta}\Big[ \nabla_{\theta} \ell_{\theta} +  \ell_{\theta}  \nabla_{\theta} \log p_{\theta}(\mathbf{h}_{\leq T}^E, \mathbf{x}_{\leq T}, \mathbf{y}_{\leq T}, \mathbf{h}_{\leq T}^D || \mathbf{u}_{\leq T}, \mathbf{v}_{\leq T}) \Big]& \nonumber \\
    % & = \mathrm{E}_{\theta}\bigg[ \nabla_{\theta} \ell_{\theta} 
    % + \ell_{\theta} \Big(  \nabla_{\theta^{E}} \log p_{\theta^E}(\mathbf{h}_{\leq T}^E, \mathbf{x}_{\leq T} || \mathbf{u}_{\leq T}) & \nonumber \\ 
    % &\qquad \qquad \qquad \quad \quad + \nabla_{\theta^{D}} \log p_{\theta^D}(\mathbf{h}_{\leq T}^D || \mathbf{y}_{\leq T}, \mathbf{v}_{\leq T}) \Big)  \bigg] & \nonumber \\ 
    = \mathrm{E}_{\theta} \bigg[ \nabla_{\theta} \ell_{\theta} 
    + \ell_{\theta} \cdot \sum_{t=1}^T \sum_{i \in \mathcal{H}^E, \mathcal{X}, \mathcal{H}^D} \nabla_{\theta_i^{S}} \ell \big( s_{i,t}, \sigma(o_{i,t})\big) \bigg].
\end{align}
% The learning signal $\ell_{\theta}$ in \eqref{eq:ls} depends on the parameters of output neurons in $\mathcal{V}$, while the causally conditioned distribution $p_{\theta}(\mathbf{h}_{\leq T}^E, \mathbf{x}_{\leq T}, \mathbf{y}_{\leq T}, \mathbf{h}_{\leq T}^D || \mathbf{u}_{\leq T}, \mathbf{v}_{\leq T})$ in \eqref{eq:caus-cond} depends on the parameters of neurons in subsets $\mathcal{H}^E, \mathcal{X}, \mathcal{H}^D$.
Accordingly, for any neuron $i \in \mathcal{H}^E, \mathcal{X}, \mathcal{H}^D$, the gradient \eqref{eq:grad-elbo} equals
\begin{align*}
    \nabla_{\theta_i^{S}} \mathcal{L}(\theta) = \mathrm{E}_{\theta}\Bigg[ \ell_{\theta} \cdot \sum_{t=1}^T \nabla_{\theta_i^{S}} \ell\big( s_{i,t}, \sigma(o_{i,t})\big) \Bigg],
\end{align*}
with $s_{i,t}$ being $h_{i,t}^E, x_{i,t},$ and $h_{i,t}^E$ respectively; while for any output neuron $i \in \mathcal{V}$, we have
\begin{align*}
    \nabla_{\theta_i^{D}} \mathcal{L}(\theta) = \mathrm{E}_{\theta}\Bigg[ \sum_{t=1}^T \nabla_{\theta_i^{D}} \ell\big( v_{i,t}, \sigma(o_{i,t})\big) \Bigg].
\end{align*}

The expectation in \eqref{eq:grad-elbo} is in practice approximated via Monte Carlo (MC) estimates by drawing a sample $\mathbf{h}_{\leq T}^E, \mathbf{x}_{\leq T}, \mathbf{y}_{\leq T}, \mathbf{h}_{\leq T}^D$ of hidden neurons from the causally conditioned distribution $p_{\theta}(\mathbf{h}_{\leq T}^E, \mathbf{x}_{\leq T}, \mathbf{y}_{\leq T}, \mathbf{h}_{\leq T}^D || \mathbf{u}_{\leq T}, \mathbf{v}_{\leq T})$ and evaluating 
\begin{align} \label{eq:grad-mc}
    \nabla_{\theta_i^{S}} \mathcal{L}(\theta) &\approx \ell_{\theta} \cdot \nabla_{\theta_i^{S}} \ell\big( s_{i,t}, \sigma(o_{i,t}) \big), ~~ i \in \mathcal{H}^E, \mathcal{X}, \mathcal{H}^D, \cr 
\nabla_{\theta_i^{D}} \mathcal{L}(\theta) &\approx \nabla_{\theta_i^{D}} \ell\big( v_{i,t}, \sigma(o_{i,t}) \big), ~~~~~~ i \in \mathcal{V}.
\end{align}
The gradients \eqref{eq:grad-mc} can be evaluated in an online manner by computing for each time-step $t$ the eligibility trace $e_{i,t}= \nabla_{\theta_i^S} \ell\big( s_{i,t}, \sigma(o_{i,t})\big)$ of each neuron $i \in \mathcal{N}^{S}$ \cite{intro_snn}. 
The resulting algorithm is detailed in Algorithm~\ref{algo1}, where we also introduce the baseline $b_{i,t}$ as a control variate to reduce the variance of estimates \cite{jang2020vowel}.

\vspace{1cm}

\begin{algorithm}[t!]
%\SetAlgoNoLine
\LinesNumbered
\KwIn{Exogeneous signal $\mathbf{u}_{\leq T}$, desired output $\mathbf{v}_{\leq T}$, and learning rates $\eta$, $\kappa_1$ and $\kappa_2$}
\KwOut{Learned model parameters $\theta$}
\vspace{0.1cm}
\hrule
\vspace{0.1cm}
{\bf initialize} parameters $\theta$ \\
\For{{\em each time $t=1, 2,\ldots$}}{
generate spike outputs from the encoding SNN 
\begin{align*}
    \mathbf{h}^{E}_{t}, \mathbf{x}_{t} \sim p_{\theta}(\mathbf{h}^{E}_{t}, \mathbf{x}_{t} | \mathbf{u}_{\leq t}, \mathbf{h}^{E}_{\leq t-1}, \mathbf{x}_{\leq t-1})
\end{align*} \\
generate outputs of the channel 
\begin{align*}
    \mathbf{y}_{t} \sim p(\mathbf{y}_{t} | \mathbf{y}_{\leq t-1}, \mathbf{x}_{t})
\end{align*} \\ 
generate spike outputs from the decoding SNN 
\begin{align*}
    \mathbf{h}^{D}_{t} \sim p_{\theta}(\mathbf{h}^{D}_{t} | \mathbf{y}_{\leq t}, \mathbf{h}^{D}_{\leq t-1}, \mathbf{v}_{\leq t})
\end{align*} \\
\smallskip
a central processor at the encoder side collects the log losses $\ell\big(v_{i,t}, \sigma(o_{i,t})\big) = \log p(v_{i, t} | o_{i, t})$ for all output neurons $i \in \mathcal{V}$ and computes the learning signal as
\begin{align}
    \ell_{t} = \kappa \ell_{t - 1} + (1 - \kappa) \sum_{i \in \mathcal{N}} \ell\big(v_{i,t}, \sigma(o_{i,t})\big))
\end{align}
\smallskip
\For{{\em each neuron $i \in \{\mathcal{H}_{E}, \mathcal{X}, \mathcal{H}_{D}, \mathcal{V}\}$}}{
update the eligibility traces $\mathbf{e}_{i,t}$ as
\begin{eqnarray} \label{eq:eligibility-update} 
    \mathbf{e}_{i,t} = \kappa \mathbf{e}_{i,t-1} + (1-\kappa) \nabla_{\theta} \log p_{\theta}(s_{i,t} | o_{i,t}),
\end{eqnarray}
where $s_{i,t}= h^{E}_{i, t}$ if $i \in \mathcal{H}_{E}$, $s_{i,t}= x_{i, t}$ if $i \in \mathcal{X}$, $s_{i,t}= h^{D}_{i, t}$ if $i \in \mathcal{H}_{D}$, or $s_{i,t}= v_{i, t}$ if $i \in \mathcal{V}$.

\smallskip
compute the time-averaged estimates 
\begin{align} \label{eq:opt-online-avg-estimate}
\langle \mathbf{e}^{2}_{i,t} \rangle &= \alpha \cdot \langle \mathbf{e}^{2}_{i,t-1} \rangle + (1-\alpha) \cdot \mathbf{e}^{2}_{i,t}, \cr 
\text{and}~~ \langle \ell_{t} \cdot \mathbf{e}^{2}_{i,t} \rangle &= \alpha \cdot \langle \ell_{t-1} \cdot \mathbf{e}^{2}_{i,t-1} \rangle + (1-\alpha) \cdot \ell_t \cdot \mathbf{e}^{2}_{i,t},
\end{align}
and estimate the baseline $\mathbf{b}_{i,t}$ as 
\begin{align} \label{eq:opt-online-baseline-estimate}
\mathbf{b}_{i,t} = \frac{ \langle \ell_t \cdot \mathbf{e}^{2}_{i,t} \rangle }{ \langle \mathbf{e}^{2}_{i,t} \rangle }
\end{align}
\smallskip
compute the time-averaged updates $\Delta_{i,t}$ as
\begin{align} \label{eq:avg-update}
\Delta_{i,t} = 
\begin{cases}
\kappa_2 \cdot \Delta_{i,t-1} & + (1-\kappa_2) \cdot \big( \ell_{t} - \mathbf{b}_{i,t} \big) \cdot \mathbf{e}_{i,t}, \\ 
& \text{if}~ i \in \{ \mathcal{H_{E}, X, H_{D}} \} \\
 \mathbf{e}_{i,t}, &\text{if}~ i \in \{ \mathcal{V}\}
\end{cases}
\end{align} \\
update the local model parameters as 
\begin{align} \label{eq:online-update}
\theta_i^{S} \leftarrow \theta_i^{S} - \eta \cdot \Delta_{i,t}
\end{align}
}
}
\caption{NeuroJSCC}
\label{algo1}
\end{algorithm}
\section{Experiments}

We now consider an example consisting of the remote detection of handwritten digits recorded by a neuromorphic camera. Code is available at \url{https://github.com/kclip}. We specifically assume that the inputs $\mathbf{u}_{\leq T}$ are selected from the MNIST-DVS dataset, which contains spike-encoded MNIST images captured with a DVS camera \cite{serrano2015poker}. We select only examples for digits ``$0$''  and ``$7$''. With the data preprocessing detailed in \cite{skatchkovsky2020flsnn}, we have $T = 80$ and $26 \times 26$ images, yielding a number $d_u = 676$ of input spiking signals. We assume a frequency-flat Gaussian channel as in \eqref{eq:channel}, with per-symbol transmission signal-to-noise (SNR) ratio defined as the average symbol power the over noise power, i.e., $\big(|| \mathbf{x}_{\leq T} ||_{1} /(d_{x}T) \big)/\sigma^2$, with $||\ \cdot \ ||_{1}$ counting the number of pulses in the argument vector. For a given SNR, we adjust the noise power $\sigma^2$ in order to ensure the same SNR level for all schemes. 

For NeuroJSCC, the encoding SNN has $d_u = 676$ exogeneous inputs, which are fed to all $d_x = r d_u$ neurons in the output layer for a rate $r$. The encoder has no hidden neurons, i.e., we have $N_{H}^{E} = 0$. The decoding SNN consists of $d_y = d_x$ inputs, $N_H^D = d_x$ internal neurons, and $d_v = 2$ output neurons, corresponding to the two classes. Directed links exist from exogeneous inputs to all neurons. For encoding and decoding SNNs, all the neurons are fully connected. Synaptic and feedback filters are selected as in \cite{pillow2008spatio}. The decision of the decoding SNN is obtained via rate decoding at the output layer by choosing the neuron with the largest number of spikes. 

\begin{figure}
\centering
\includegraphics[scale=0.315]{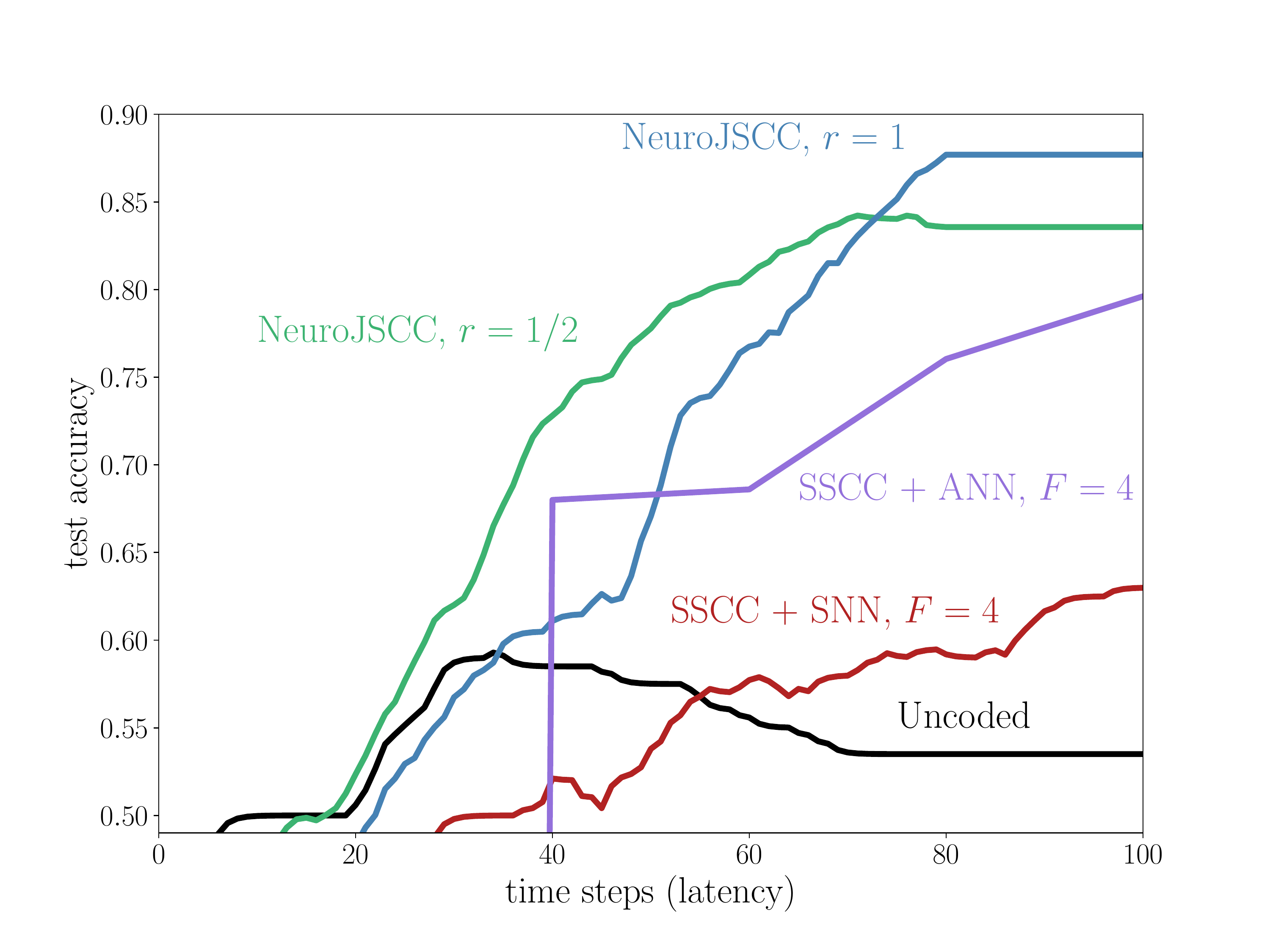}
\caption{Test accuracy as a function of the observation time-steps during inference for Uncoded transmission and NeuroJSCC schemes (SNR$=-6$ dB).}
\label{fig:acc_per_timestep}
\end{figure}

We compare NeuroJSCC to two benchmark schemes.
\begin{enumerate}
    \item \textit{Uncoded transmission}: The observation $\mathbf{u}_t$ is directly transmitted through the Gaussian channel \eqref{eq:channel} using On-Off Shift Keying (OOK), requiring rate $r = 1$. At the receiver side, the noisy samples are classified using the described decoding SNN with $N_{H}^{D} = 256$ hidden neurons, which is trained using Algorithm~\ref{algo1} for the special case in which the encoding SNN is not present (see \cite{intro_snn}). The element-wise binary quantizer in \eqref{eq:channel} is chosen as
    \begin{align} \label{eq:threshold}
    Q(x) = 
    \begin{cases}
    1 & \text{if}~ x \geq 0.5 \\
    0 &\text{if}~ x < 0.5
    \end{cases}.
    \end{align}
    We note that adaptive techniques to select the threshold may improve the performance of the uncoded scheme \cite{cha2020noncoherent}.

    \item \textit{Separate Source-Channel Coding (SSCC)}: The encoder applies state-of-the-art quantization based on the Vector-Quantization Variational Autoencoder (VQ-VAE) scheme \cite{oord2017discretevae}, followed by LDPC encoding. Compression and LDPC code rate are chosen so as to obtain a rate $r = 1$, by choosing a compression rate of $2$ and a channel encoding rate of $1/2$. The scheme is applied separately to each one of $F$ frames of size $\left \lceil{T/F}\right \rceil$ samples of the input $\mathbf{u}_{\leq T}$. At the decoder side, each frame is decoded using the Belief Propagation algorithm, decompressed using VQ-VAE decoding, and then classified. We consider two different classifiers, namely traditional ANN \cite{deng2019comparison} and SNN \cite{intro_snn}. VQ-VAE and classifiers are trained trained separately using the original, noiseless, data. The SNN is defined and trained as for the uncoded scheme, while the ANN has a single hidden layer comprised of $256$ neurons. For the implementation with an ANN, we allow the output of the VQ-VAE to be real, instead of binary as in the case of the implementation with an SNN as classifier. 
\end{enumerate}

%  As mentioned by the authors of  \cite{choi2019jssc}, we also tried to implement several internet-standard schemes for compression (e.g., LZMA, ZLIB, ...) but the noise levels at which we operate do not allow for a decoding of the received corrupted data. 

In Fig.~\ref{fig:acc_per_train_ite}, we start by demonstrating the evolution of the test accuracy of NeuroJSCC over the training iterations with rates $r = 0.5, 1, \text{and } 1.5$, at different SNR levels. NeuroJSCC is seen to approach the performance of ideal transmission even at an SNR of 0 dB with a rate $r = 1$, and the performance is robust even at degraded SNR levels. For an SNR as low $-6$ dB, the test accuracy remains higher than $88 \%$ for $r = 1$.
We also note that increasing the transmission rate for NeuroJSCC is not necessarily beneficial, which is likely due to overparametrization and resulting overfitting. 

We then evaluate the performance during inference in terms of \textit{time to accuracy} \cite{davies2019spikemark}, i.e., in terms of test accuracy as a function of the number of observed time samples. We note that NeuroJSCC and Uncoded transmission have zero latency, in the sense that a time sample is directly transmitted to the receiver without the need to form frames. In contrast, with SSCC, a decision can only be made after transmission and processing of a frame. To allow for a fair comparison with the other schemes, we assume that the recording and transmission of a frame is achieved after $\lceil T/F \rceil$ time samples. Its classification with the ANN takes $\lceil T/F \rceil$ more time steps. The number of time samples can hence be interpreted as a measure of latency. As the SNNs in NeuroJSCC performs online transmission and classification, the figure demonstrates a graceful trade-off between the number of processed samples and the classification performance. It also highlights the achievable trade-offs between accuracy and latency.

\begin{figure}
\centering
\includegraphics[scale=0.315]{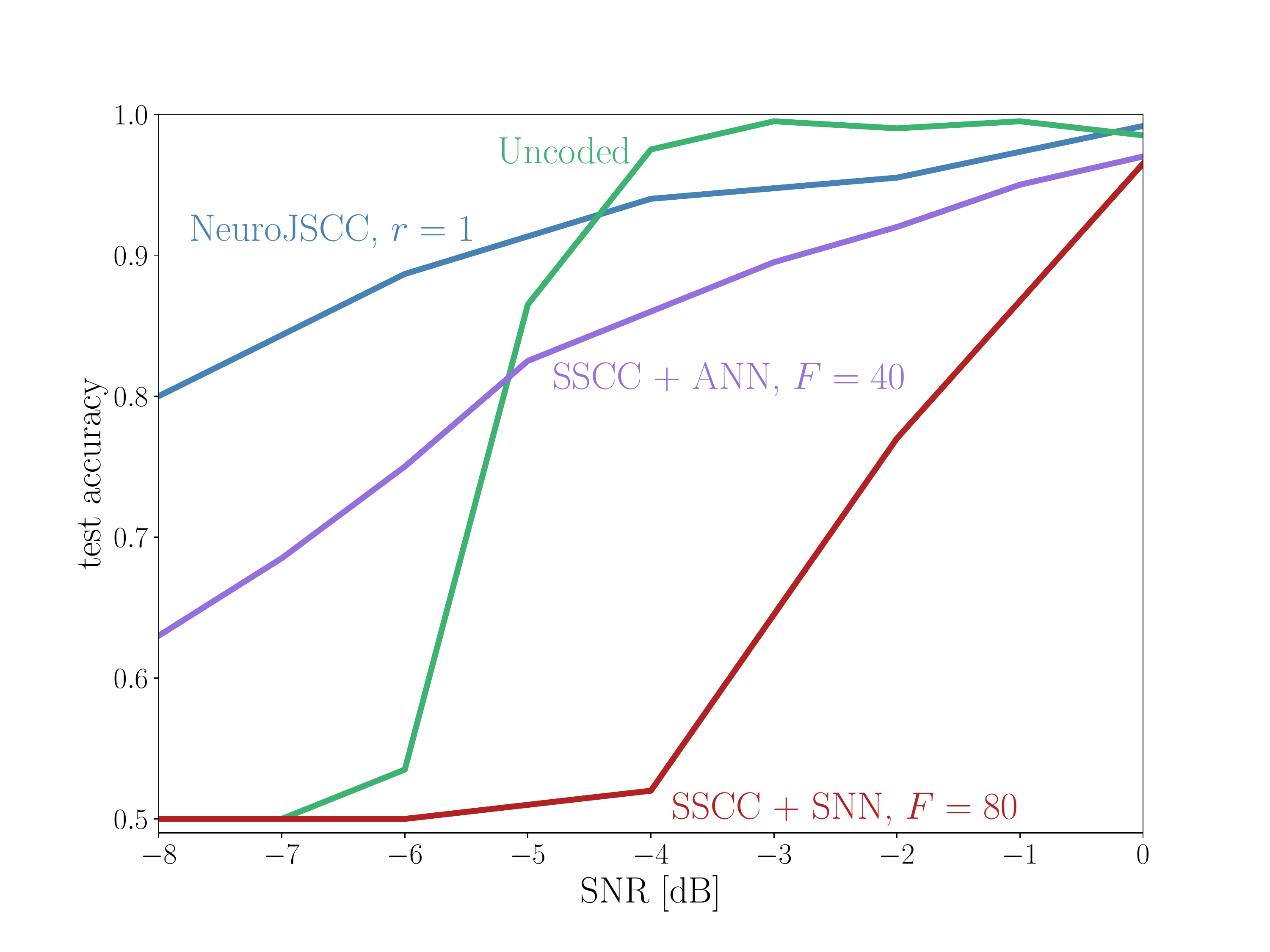}
\caption{Test accuracy at different SNR levels for the Uncoded, Separate SCC and NeuroJSCC schemes.}
\label{fig:acc_per_snr}
\end{figure}

In Fig. \ref{fig:acc_per_snr}, we evaluate the test accuracies at convergence obtained for different levels of SNR by considering also SSCC. The accuracy of Uncoded transmission drops sharply at sufficiently low SNR levels. In contrast, NeuroJSCC maintains a test accuracy of $80 \%$, even at an SNR level as low as $- 8$ dB. Separate SCC with an SNN as classifier suffers the most from the degradation of the SNR. Using an ANN proves more robust to low SNR levels, since an ANN can benefit from the non binary outputs of the VQ-VAE decoder without further loss of information due to binary quantization. 

% Note that the SNR levels shown in these figures may appear very low. Using binary signals, the power of the noise depends on the sparsity of the spikes. To provide the reader with a clearer understanding of the orders of magnitude, we provide in Table ~\ref{tab:ber_per_snr} the average BER when transmitting MNIST-DVS images via a Gaussian channel. 

% \begin{table}[h!]
% \centering
% \begin{tabular}{ | m{0.6cm} || m{0.6cm}| m{0.6cm} | m{0.6cm} | m{0.6cm}| m{0.6cm} | m{0.6cm} |m{0.6cm} |m{0.6cm} |m{0.6cm} | } 
%  \hline
%  SNR & -2 & -3 & -4 & -5 & -6 & -7 & -8 \\
%  \hline
%  BER & 0.97 & 0.95 & 0.91 & 0.86 & 0.81 & 0.76 & 0.71 \\
% \hline
% \end{tabular}
%   \caption{Illustration of the BER obtained for the transmission of MNIST-DVS samples via a Gaussian channel.}
% \label{tab:ber_per_snr}
% \end{table}

\begin{figure}
\centering
\includegraphics[scale=0.315]{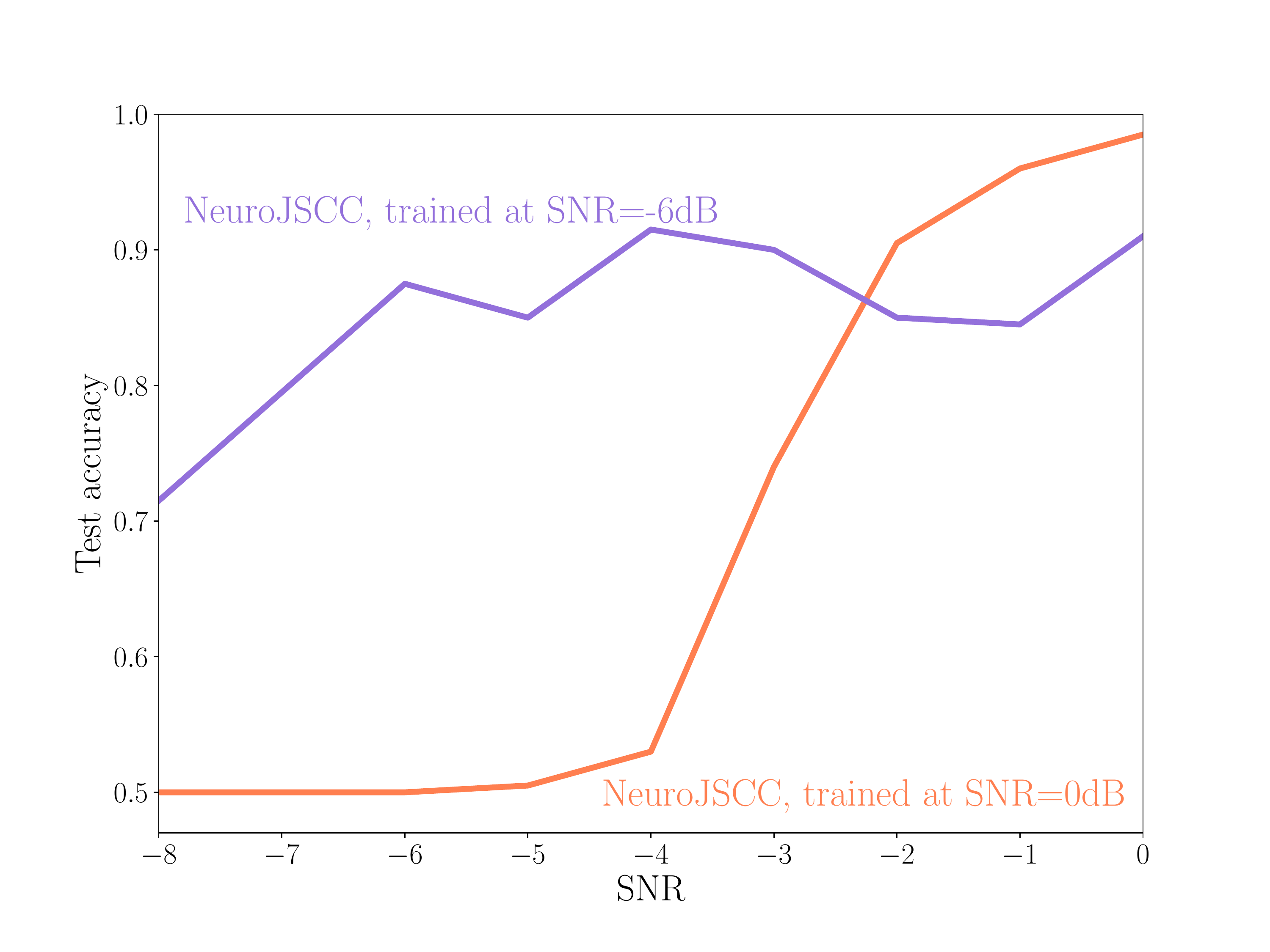}
\caption{Test accuracy of NeuroJSCC trained at various SNR levels as a function of the SNR.}
\label{fig:gen_acc_per_snr}
\end{figure}

The results plotted in Fig.~\ref{fig:acc_per_snr} were obtained for models trained separately at each SNR level. Separate training is computationally expensive. To evaluate the impact of a mismatch between SNR conditions between training and testing, we train NeuroJSCC at a single SNR, and measure its test accuracy obtained at different SNR levels. In Fig.~\ref{fig:gen_acc_per_snr}, we observe that NeuroJSCC trained at a SNR of $-6$ dB performs well also at higher SNRs, suggesting the robustness of the scheme to an SNR mismatch.

% \note{I had the idea to complete this section by training the system by randomly selecting the SNR in [0, -10] for each sample; I've launched the experiments and should have the results in the next days.}

% To improve upon this result, we train the system by randomly selecting an SNR in the range $[0, 30]$ dB for each example during training. The results are plotted in Fig.~6.b. We note that \note{...}

\section{Conclusions}

In this work, we have introduced a neuromorphic sensing, communication, and remote inference system, that is based on neuromorphic sensing, Spiking Neural Networks (SNNs), and Impulse Radio (IR). The system is trained end-to-end using a probabilistic autoencoder formulation that yields a local learning rule with global feedback. Thanks to the sparsity of computing and communication, which directly reflects the sparsity in the input spiking data, the scheme is seen to be extremely efficient, yielding high accuracy, larger than $80 \%$, on standard neuromorphic data sets, even at SNRs as low as $- 8$ dB. Future work may include the performance analysis over multipath fading channels and the study of multi-access systems. 
\newpage
\appendices

\newpage
\bibliographystyle{IEEEtran}
\bibliography{biblio}

% that's all folks
\end{document}